# Discretization-based ensemble model for robust learning in IoT

Anahita Namvar[1,2], Chandra Thapa[2], Salil S. Kanhere[1]

[1] UNSW Sydney, NSW 2052 Australia
[2] Data61 Marsfield, NSW 2122 Australia
a.namvar@student.unsw.edu.au, chandra.thapa@data61.csiro.au,
salil.kanhere@unsw.edu.au

**Abstract.** IoT device identification is the process of recognizing and verifying connected IoT devices to the network. This is an essential process for ensuring that only authorized devices can access the network, and it is necessary for network management and maintenance. In recent years, machine learning models have been used widely for automating the process of identifying devices in the network. However, these models are vulnerable to adversarial attacks that can compromise their accuracy and effectiveness. To better secure device identification models, discretization techniques enable reduction in the sensitivity of machine learning models to adversarial attacks contributing to the stability and reliability of the model. On the other hand, Ensemble methods combine multiple heterogeneous models to reduce the impact of remaining noise or errors in the model. Therefore, in this paper, we integrate discretization techniques and ensemble methods and examine it on model robustness against adversarial attacks. In other words, we propose a discretization-based ensemble stacking technique to improve the security of our ML models. We evaluate the performance of different ML-based IoT device identification models against white box and black box attacks using a real-world dataset comprised of network traffic from 28 IoT devices. We demonstrate that the proposed method enables robustness to the models for IoT device identification.

## 1    Introduction

The Internet of Things (IoT) has revolutionized the way we interact with our surroundings, connecting a wide range of objects, from household appliances to smartphones and automobiles. However, this connectivity has also introduced several challenges for organizations, particularly with regard to managing access and preventing potentially insecure IoT devices from connecting to their network [1]. To address this challenge, accurate and fast methods are needed to effectively manage the visibility of IoT devices on the network and distinguish compromised devices from legitimate ones.

Machine learning models have emerged as a promising approach for IoT device identification, as they can learn to recognize the unique behavior patterns of each device on the network [2]. However, these models are susceptible to various types of noise and adversarial attacks that can potentially affect their accuracy and performance [3-5]. Hence, the question arises as to how these models can be made more robust to overcome such challenges. In order to improve the robustness of machine learning models, researchers often employ discretization techniques to reduce the impact of noise on the model's performance. This technique involves converting continuous features into a finite set of values, which makes the model less sensitive to small variations in the input data [6]. Additionally, ensemble methods can be used to combine multiple models in order to reduce the impact of any remaining noise or errors [7]. However, in this paper, we go beyond these techniques and propose a novel approach that leverages the benefits of both discretization and ensemble methods in a unique way to improve the robustness of IoT device identification classifiers. Specifically, we propose a discretization-based ensemble stacking technique that merges multiple models trained on different subsets of the discretized data with different hyperparameters. By doing so, we can achieve better performance than either discretization or ensemble methods alone. Thus, in this paper, we explore the individual benefits of these techniques and propose a new methodology that merges them in a novel way to improve the robustness of IoT device identification classifiers in an adversarial environment.

We summarized our contribution in the following:

1. We investigate the effectiveness of different discretization methods as a defense mechanism for IoT device identification classifiers. Our goal is to improve the security of these classifiers against white box and black box attacks.

2. We demonstrate the most effective discretization methods through our experiments. We show that these methods can provide an additional layer of protection against different adversarial attacks on IoT device identification classifiers.

3. We propose a novel discretization-based ensemble defense methodology that utilizes discretization and stack ensemble. We evaluate the effectiveness of our proposed method on the state-of-the-art IoT device identification classifiers and against various White box black box adversarial attacks using a real-world dataset comprised of network traffic from 28 IoT devices. The findings present exceptional resilience of robust models against adversarial attacks while maintaining high accuracy in a clean environment.

The rest of this paper is organized as follows: Section 2 provides background on discretization and ensemble-based robustness, Section 3 presents the details of our research methodology, and Section 4 provides our experiments and results. The relevant related works are overviewed in Section 5, and Section 6 concludes the paper.

An adversarial attack is a deliberate attempt to deceive or confuse the classifier using small changes to the input data.

## 2  Background and related works

### 2.1  IoT device Classification and adversarial attacks

Machine learning-based IoT device identification models emerge for the effective and secure management of IoT networks. It allows network administrators to monitor device behavior, detect and mitigate threats, and allocate resources efficiently. Recently it has been shown that machine learning models are vulnerable to adversarial attacks. The adversarial attack is a malicious behavior that intentionally adds subtle perturbation to input data to fool the machine learning model ([8, 9]).

The typical methods to generate adversarial examples include Fast Gradient Sign Method (FGSM), Basic Iterative Method (BIM), and Jacobian Saliency map Attacks (JSMA) [10-12]. The Fast Gradient Sign Method (FGSM) attack is a classical method that works by manipulating the gradients of a machine learning model to generate adversarial examples [9]. This attack involves taking the gradient of the loss function with respect to the input and then perturbing the input in the direction that maximizes the loss. The magnitude of the perturbation is controlled by a parameter called epsilon value. By adding a small perturbation to the original input data, an attacker can cause the machine learning model to misclassify the input data. BIM is another attack, and it works by applying the FGSM attack multiple times iteratively [13]. JSMA is an adversarial attack to generate imperceptible adversarial examples to fool machine learning models [14]. This attack relies on perturbing the input data in a way that maximizes a specific target class while minimizing the changes to the original input.

The vulnerability of machine learning models to adversarial attacks is mostly addressed in the image domain, and less amount of research has been done on the security of machine learning base device identification models. Namvar, et al. [5] presented the vulnerability of device identification models, including Random Forest (RF), Logistic Regression (LR), Feed Forward (FF), and Decision Tree (DT) to adversarial attacks. However, they did not provide any solution to increase the models' robustness against attacks. Bao, et al. [15] is another study that provides a robust deep-learning model for device identification. Singh and Sikdar [16] investigated the effectiveness of generating attacks against a deep learning-based appliance classification model using smart meter data. The study found that adversarial attacks can significantly degrade the accuracy of the model. To investigate the detectability of attacks in the smart home domain, the authors proposed a visualization method that compares the distribution of true data points and adversarial data points. The findings of the study highlight the need for effective defense mechanisms to counter adversarial attacks on deep learning-based appliance classification models in the smart home domain. Meidan, et al. [17] proposed random forest supervised machine learning model to detect suspicious IoT devices connected to organizational networks. The results showed that the algorithm correctly classified white-listed device types in 99% of cases and detected non-white-listed device types in 96% of test cases. The authors presented the resilience of their method to adversarial attacks. Kotak and Elovici [18] propose a new approach for generating real-time adversarial examples using heatmaps generated by CAM and Grad-CAM++ that can be applied in the computer network domain.  Authors revealed that in many cases, an adversarial example created using a heatmap could deceive the payload-based deep IoT device identification solution with up to 100% accuracy.

Luo, et al. [19] proposed an adversarial machine learning-based partial-model attack strategy, which mainly sits in the data-collecting and aggregating stage of IoT systems. results demonstrate that the machine learning engine of the IoT system is highly susceptible to attacks. These attacks exploit the uncertainty in the performance of IoT devices or the communication channel, and the consequences of such attacks can result in significant disruption to the overall operations of the IoT system. The authors discussed the necessity of countering these adversarial attacks by providing appropriate defence mechanisms and leave it to their future work.

## 2.2 Discretization-based ensemble as a defense strategy.

Discretization, also known as quantization or binning, is the process of converting continuous input variables to discrete values by dividing the input range into a fixed number of bins or intervals [20]. Data discretization is a widely used technique in machine learning, and it has a critical role in data cleansing and preprocessing. The process typically works as follows: The model analyses the input data and determines which variables have a large number of unique values or are highly correlated with input. Then the bin boundaries are defined by specifying the number of bins to use and determining the range of values for each bin. Once the bin boundaries have been defined, each input value is assigned to the appropriate bin based on its value. Then the bin boundaries are encoded using one hot encoding. Finally, the original input data is replaced with the discretized versions in the model.

Discretization has been introduced for secure deep learning in image datasets [21]. Sharmin, et al. [22] improved the adversarial robustness of the spiking neural network classifier by input discretization through poison encoding. Panda, et al. [6] showed discretizing the input space, which allowed pixel levels from 256 values or 8bit to 4 values or 2bit extensively improve the adversarial robustness of the DLNs for a substantial range of perturbations for minimal loss in test accuracy. Lal, et al. [23] demonstrate cluster structure discretization proposing a robust defensive model for retinal fundus images for Diabetic Retinopathy. Feng, et al. [24] proposed a content-adaptive pixel discretization defense called Essential Features, which discretizes the image to a per-image adaptive codebook to reduce the color space. They applied PGD and BPDA white box attacks and Square Attack as a black box attack. Until recently, discretization robustness was mostly studied in computer vision. However, only one research presented discretization as a defense mechanism in protecting tabular data from adversarial attack [25].

There are several methods for discretization [26]:

**1 - Equal Width (EW):** Equal width method is the most used algorithm that divides the range of a feature into a fixed number of bins of equal size.

**2- Equal Frequency (EF):** Equal frequency algorithm divides the range of a feature into bins such that each bin contains roughly the same number of observations.

**3- Decision Tree-based Binning (DTB):** Decision tree-based binning is the decision tree-based method that divides the range of a feature into bins.

**4- Entropy-based Binning (EBB):** Entropy-based binning method uses the concept of entropy from information theory to divide the range of a feature into bins.

**5- Minimum descriptive length (MDL):** Minimum descriptive length method finds the best partitioning of the feature space into bins. This is achieved by minimizing the sum of the description lengths of the bins and the description lengths of the data within the bins. MDL is computationally expensive as it requires finding the best partitioning of the feature space, which can be a difficult optimization problem.

Fig.1 illustrates a simplified view of data discretization and shows how it improves robust machine learning against adversarial attacks. The black circle in Fig1.A presents the adversarial sample that is maliciously tampered with by an attacker. The attacker adds slight perturbation $\varepsilon$ to the input data, which leads to misclassification. However, let us suppose that we have defined bin boundaries $(C_1, C_2, ..., C_n)$ after applying a discretization algorithm. The bin boundaries can be seen in Fig1. B. After defining bin boundaries, each input data is assigned to the appropriate bin. Finally, the original input data is replaced with the discretized versions in the model. Fig1.C depicts the discretized samples. Hence, the small changes in the input data cannot fool the model. For example, both X1 and $\hat{X}_1$ will be categorized in bin 3, and adversarial attacks cannot simply misclassify input data.

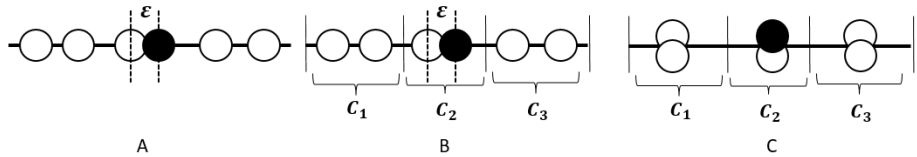

**Fig. 1.** A simple view of discretization for improving robustness.
A. Data before discretization. Continuous variables, and black circle presents an adversarial sample.
B. Data is discretized into three equal ranges. All data in the same category have been mapped to an equal value.
C. Data after discretization.

Discretization reduces the number of possible inputs, making it harder for an attacker to find inputs that cause the model to make a mistake [27]. Discretization also reduces model sensitivity to small changes in the input, which can make it less susceptible to adversarial examples [25]. Discretization can increase interpretability which can improve model robustness. Finally, discretization can smooth the decision surface, which makes the model robust to small changes in the input data [28].

Ensemble robustness is another method that can increase the security of machine learning models against adversarial attacks[29]. Ensemble methods combine the predictions from multiple models to seek better predictive performance, so it is less likely that attacker can find a single small perturbation that can fool all the models. Each model in the ensemble may make different errors, so an adversary example that successfully full one model may not attack all models in the ensemble. Mohanty, et al. [7] showed that averaging the output prediction of multiple models can increase adversarial robustness. Tramr, et al. [30] showed that training the ensemble model on adversarial examples can improve the model's robustness to attacks. Pang [31] showed that the diversity of models in an ensemble can make it difficult for attackers to fool all the models in the ensemble. Grefenstette, et al. [32] demonstrated that adversarial training of ensemble presents better robustness than adversarial training multiple machine learning models and takes an ensemble of these models. Li and Li [33] proposed an ensemble of deep neural networks to enhance robustness against adversarial attacks. Dabouei, et al. [34] proposed regularization as a vigorous approach to encourage the model to be robust to adversarial examples.

In this study, we illustrate the idea of using an ensemble of different robust discretized classifiers. Discretization can reduce the number of inputs and make the model less sensitive to small changes, while ensemble methods can improve the model's predictive performance and make it harder for an attacker to find a single weakness. we selected Minimum descriptive length, Equal Frequency, Equal Width, and Entropy-based Binning techniques because of their simplicity, frequency of use, preventing information leakage and preventing information loss.

## 3    Methodology

This section discusses the proposed methodology for the robust learning-based IoT device identification mechanism that can improve the resilience of models in an adversarial environment. The first step is to discretize the input data using three different methods of EW, MDL, and EBD. The next step is training an IoT device identification classifier separately on an original dataset for each discretization method. This means we will have three IoT device identification classifiers, each trained on a different discretization method. These three models create the base learners for stack ensemble classifies. The final step is to combine the three sets of predictions using a stack ensemble to obtain one final prediction. We then test the model using imperceptible adversarial examples to investigate the effectiveness of our robustness methodology in improving the accuracy of machine learning-based IoT device classifiers when exposed to adversarial attacks such as FGSM, JSMA, and BIM.

The proposed methodology can be integrated into a larger network security system, where it can serve as an additional layer of defense against adversarial attacks. The algorithm can be deployed on edge devices or gateways within the network to identify IoT devices and detect any malicious activity. Furthermore, the methodology can be combined with other security mechanisms, such as firewalls and intrusion detection systems, to provide a more comprehensive approach to network security.

Fig1 Illustrated the overall methodology of the proposed defense method. The details of the proposed methodology are discussed in the following sections.

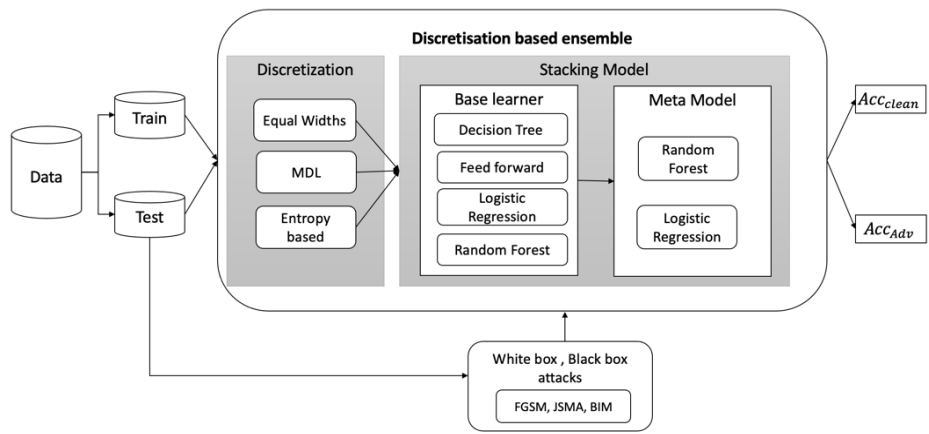

**Fig. 2.** An overview of discretization-based ensemble robustness.

### 3.1 Input Discretization

First, we discretized input features into categorical features; we used four state-of-the-art discretization methods, EF, EW, MDLP, and entropy-based discretization. These methodologies were chosen due to their ability to effectively discretize continuous data. EF approach relies on dividing the range of the continuous variable into equal-sized intervals. In the EW approach, every attribute is divided into intervals of equal width. MDL relies on defining a set of possible discretization schemes, then calculate the description length for each discretization scheme. The discretization scheme with the shortest description length will be selected for converting the continuous variable into categorical variables. Entropy-based binning partitions the feature space based on the information gain. The method aims to minimize the overlap between different classes, which can make the model more robust to adversarial examples. We train our classifiers with discretized features to evaluate the defense capability of models.

### 3.2 Ensemble

The ensemble method addressed in this study is the stacking ensemble. we address EW, MDL, and entropy discretization techniques since they lead to higher performance on robust accuracy. The stack ensemble model involves two levels of machine learning models: level 0 and level 1. The former consists of single models trained on discretized input features, while the latter is limited to only one meta-model of either RF or LR that combine the outcome of discretized single models.

### 3.3 Adversarial attack

In an IoT network, adversarial attacks pose a significant threat to the security and privacy of devices and their users. For instance, attackers may deploy malware on IoT devices and manipulate their traffic to gain control and generate adversarial examples. In this study, the most state-of-the-art adversarial attacks, including FGSM, BIM, and JSMA, were employed under two scenarios of a white box and a black box to evaluate the robustness of machine learning-based IoT device identification models. These attacks can be generated by adding imperceptible perturbations to the input data, which can evade the machine learning models and cause misclassification of IoT devices [5]. For example, an attacker could generate adversarial examples to make a device appear as a legitimate device to avoid detection while it is actually malicious and performing harmful actions on the network. Therefore, it is essential to develop robust machine learning-based IoT device identification mechanisms that can withstand adversarial attacks and improve the overall security of IoT networks.

## 4 Experimental results

### 4.1 Dataset

we utilized a comprehensive dataset obtained by Sivanathan, et al. [35]. The dataset comprises network traffic information derived from a total of 28 IoT devices. Containing an array of device types, such as cameras, lights, plugs, motion sensors, appliances, and health monitors. The dataset includes 7 key attributes: traffic flow, traffic flow, flow volume, flow duration, sleep time, NTP interval, DNS interval, and domain count.

### 4.2 Adversarial attack setting and evaluation metric:

To test the robustness of the proposed defence mechanism, our work employs three states of the art adversarial attacks, namely FGSM, BIM, and JSMA, in two scenarios of white box and black box attack. Our methodology for black box attacks involves three state-of-the-art adversarial attacks: FGSM, BIM, and JSMA. We generate adversarial perturbations using substitute models of LR and FF. To simplify our notation, we use the following abbreviations: FGSM LR for FGSM attack with LR as a substitute model, FGSM FF for FGSM attack with FF as a substitute model, and so on. A complete list of the different combinations of attack and substitute models can be found in Table 2. Such black box attack methods are likely to effectively fool the model due to the transferability property [14, 36]. For the white box setting, the attacker has full knowledge of the learning model, model parameters, and data. This study uses the default setting in IBM's Adversarial Robustness Toolbox [37] for crafting adversarial examples. The FGSM attack parameter $\epsilon$ is set to o.o1. For a BIM attack, parameters $\epsilon$ and $\alpha$ are set to 0.01, and 0.001, respectively. Finally, for the JSMA attack, the parameters, Theta, and gamma are set to 0.01, 0.4, respectively. White box attack setup is used for differentiable classifiers of LR and FF. On the other hand, black box attacks are crafted against both differentiable and nondifferentiable models, including LR, FF, RD, and DT.

Regarding the evaluation metric, we define the robustness measure as the percentage drop in model accuracy when comparing the model's performance in a clean and adversarial environment. Equation X illustrates the Robustness measure (RM). Lower RM presents better robustness.

$$RM = \frac{\|Acc_{Clean} - Acc_{adv}\|_1}{|Acc_{Clean}|}$$

### 4.3 Discretization

In this study, we have four IoT device classification models: Random Forest, Decision Tree, Logistic Regression, and Feedforward Neural Network. We evaluate the models' robustness under EF, EW, MDL, and entropy-based input discretization methodologies considering both white box and black box attacks. The experiments have been run in Python 3.8 on an Intel® Core i5 2.3 GHz CPU, 64 GB of RAM, and running on MAC OS.

First, we partitioned each feature with different interval numbers ranging from 1 to 100 and compared the robustness of classifiers for each interval number. The goal is to find the optimum number of intervals that can provide better robustness. To do this, the approach tests each number of intervals by partitioning features and calculating classification accuracy in both clean and adversarial environments. The best cut point is then selected as the one that leads to the least drop in the model accuracy under adversarial attack while also maintaining high performance on clean data.

Table 1 shows the accuracy of models when the models are not exposed to adversarial manipulation. Original accuracy presents the performance of models in a clean environment where there is no discretization or ensemble mechanism involved. We use this information as a baseline for evaluating the performance of robustness mecha-nisms. According to Table 1, it is apparent that the EF discretization method has led to a significant decrease in the accuracy of the models. The original models have high accuracy, while the model built using the EF discretization method has much lower accuracy. This indicates that the discretization process has negatively impacted the ability of the models to accurately predict outcomes. This is mainly due to the distribution of the data, as the model cannot discretize the data into bins with equal frequency when the frequency in some bins is very high. The maximum number of bins for equal frequency is 17, and the model is unable to increase this limit. Therefore, since the EF approach performs poorly in a clean environment, we will not evaluate its robustness capability in the next steps, and we present implementation results for the remaining methods, namely EW, MDL, EBD, EN-LR, and EN-RF. EN-LR, and EN-RF are our proposed discretization stack ensemble methodology. From Table 1 it can be concluded that discretized-based ensemble methodology can accurately identify IoT devices in a clean environment. We will evaluate the performance of these models in an adversarial environment.

*Table 1 – Models' performance in clean environment*

| Model | original | EF | EW | MDL | EBD | EN-LR | EN-RF |
|-------|----------|--------|--------|--------|--------|--------|--------|
| DT | 99.41% | 50.67% | 93.61% | 84.15% | 85.42% | 96.27% | 96.30% |
| RF | 99.90% | 50.91% | 99.71% | 99.88% | 99.80% | 99.70% | 99.71% |
| LR | 92.99% | 32.28% | 99.68% | 99.88% | 99.82% | 99.67% | 99.59% |
| FF | 92.07% | 50.26% | 99.21% | 99.80% | 99.71% | 99.60% | 99.55% |

Table 2 presents the accuracy of different discretized models when subjected to different black box attacks.

*Table 1 - black box attacks*

| Model | Black box attack | EW | | MDL | | EBD | | No Discretisation | |
|---|---|---|---|---|---|---|---|---|---|
| | | Adv_acc | RM | Adv_acc | RM | Adv_acc | RM | Adv_acc | RM |
| **Decision Tree** | FGSM LR | 86.84% | 7.23% | 74.51% | 11.46% | 73.49% | 13.97% | 75.78% | 23.77% |
| | JSMA LR | 85.36% | 8.81% | 78.35% | 6.89% | 78.41% | 8.21% | 75.67% | 23.88% |
| | BIM LR | 87.10% | 6.95% | 74.53% | 11.43% | 73.69% | 13.73% | 75.86% | 23.69% |
| | JSMA FF | 81.34% | 13.11% | 71.18% | 15.41% | 74.06% | 13.30% | 74.37% | 25.19% |
| | FGSM FF | 87.27% | 6.77% | 74.35% | 11.65% | 75.20% | 11.96% | 78.11% | 21.43% |
| | BIM FF | 86.28% | 7.83% | 71.55% | 14.97% | 73.75% | 13.66% | 76.61% | 22.94% |
| **Random Forest** | FGSM LR | 96.87% | 2.85% | 93.83% | 6.06% | 94.91% | 4.90% | 64.82% | 35.12% |
| | JSMA LR | 98.57% | 1.14% | 98.49% | 1.39% | 98.69% | 1.11% | 73.92% | 26.01% |
| | BIM LR | 97.04% | 2.68% | 96.34% | 3.54% | 96.36% | 3.45% | 66.11% | 33.82% |
| | JSMA FF | 97.18% | 2.54% | 94.77% | 5.12% | 96.22% | 3.59% | 71.52% | 28.41% |
| | FGSM FF | 96.24% | 3.48% | 97.91% | 1.97% | 96.75% | 3.06% | 73.88% | 26.05% |
| | BIM FF | 96.42% | 3.30% | 94.58% | 5.31% | 93.25% | 6.56% | 70.44% | 29.49% |
| **LR** | JSMA FF | 95.91% | 3.78% | 92.91% | 6.98% | 92.89% | 6.94% | 65.21% | 29.87% |
| | FGSM FF | 94.77% | 4.93% | 96.69% | 3.19% | 95.20% | 4.63% | 78.13% | 15.98% |
| | BIM FF | 95.24% | 4.45% | 93.72% | 6.17% | 92.29% | 7.54% | 73.83% | 20.60% |
| **FF** | FGSM LR | 92.54% | 3.81% | 91.97% | 7.85% | 87.57% | 12.18% | 79.42% | 13.74% |
| | JSMA LR | 90.97% | 5.45% | 96.36% | 3.45% | 92.07% | 7.66% | 65.76% | 28.58% |
| | BIM LR | 92.52% | 3.84% | 91.95% | 7.87% | 87.20% | 12.55% | 78.50% | 14.74% |

The Adv_acc column indicates the accuracy of models against black-box attacks, while RM is the robustness measure and presents the percentage of drop in model accuracy comparing the model performance in a clean and adversarial environment. Most of the discretized models exhibit a high level of adversarial robustness, as evidenced by their accuracy in an adversarial environment, compared to their lower accuracy when no discretization mechanism is used. EW is proving to be a robust technique for all classifiers, as it achieves high accuracy in adversarial environments and exhibits less degradation in accuracy when exposed to adversarial attacks (low RM measure). for example, in the Decision Tree model, the Adv_acc for FGSM LR attack is 86.84% in the EW model, whereas it is only 75.78% without discretization. Similarly, in the Random Forest model, the Adv_acc for JSMA LR attack is 98.57% in the EW model, whereas it is only 73.92% in the no discretization model. This can be attributed to its ability to handle data with varying distributions and to its effectiveness in reducing the impact of outliers. EW discretizes the variable into a smaller number of categories; hence, the noise and variability can be reduced, and the model can learn more general and robust patterns. Additionally, equal-width discretization is a simple and easy-to-implement technique that requires minimal computational resources. Overall, the strong performance of equal-width discretization suggests that it is a promising approach for achieving adversarial robustness in this domain.

Results demonstrate the effectiveness of the MDL methodology in increasing the resilience of random forest, feedforward, and logistic regression models to the black box. The MDL methodology achieves this by minimizing the description length of the model, which encourages the selection of simpler and more interpretable models that are less prone to overfitting. In addition to improving model robustness, the MDL methodology provides a principled approach for selecting the best model among a set of candidates based on the tradeoff between model accuracy and complexity. Overall, the MDL methodology is a promising approach for achieving robustness in IoT device classification. The EBD discretization method has emerged as an effective way to increase the robustness of machine learning models to black box attacks, except for the decision tree model. For example, in the case of the random forest model, the EBD approach improved the adversarial accuracy against the JSMA LR attack from 73.92% to 98.69% and against the FGSM FF attack from 73.88% to 96.75%. Overall, the effectiveness of the EBD approach in improving the robustness of models against black-box attacks appears to be model-dependent. Further analysis to understand the factors that affect the effectiveness of EBD will be left for future research.

*Table 2- white box attacks*

| Model | White box attack | EW | | MDL | | EBD | | No Discretisation | |
|---|---|---|---|---|---|---|---|---|---|
| | | Adv_acc | RM | Adv_acc | RM | Adv_acc | RM | Adv_acc | RM |
| LR | FGSM | 94.79% | 4.91% | 96.50% | 3.38% | 94.44% | 5.39% | 60.10% | 35.37% |
| | JSMA | 96.50% | 3.19% | 98.22% | 1.66% | 97.20% | 2.62% | 64.82% | 30.29% |
| | BIM | 94.83% | 4.87% | 96.46% | 3.42% | 94.32% | 5.51% | 64.98% | 30.12% |
| FF | FGSM | 93.30% | 5.96% | 93.01% | 6.80% | 90.39% | 9.35% | 76.00% | 17.45% |
| | JSMA | 88.12% | 11.18% | 93.17% | 6.64% | 85.24% | 14.51% | 62.29% | 32.34% |
| | BIM | 94.89% | 4.35% | 91.23% | 8.59% | 87.84% | 11.90% | 80.29% | 12.79% |

Table 3 illustrates the adversarial robustness of discretized models to white-box attacks. All classifiers without input discretization are vulnerable to such attacks. In contrast, IoT device classifiers trained on discretized input data generated by EW, MDL, EBD show less vulnerability to adversarial attacks as their Adv_acc is much greater than that of models without discretization. Also, discretized models present lower RM compared to models with no discretization, which confirms the effectiveness of input discretization with EW, MDL, EBD for creating a robust IoT device identification model in a white box scenario.

### 4.4 Discretization-based ensemble defense mechanism

Two discretisation-based ensembles of EN-LR, EN-RF were built in this study. The ensemble method employed was a stacking method using the Scikit-learn Python library. Both models are based on ensembles of single models trained on discretised input features as their base classifiers. The main difference between EN-LR and EN-RF is their choice of meta-model. EN-LR uses LR as its meta model, while EN-RF uses RF as its meta-model. For example, to build EN_LR for DT model, The DT with EW discretisation, DT with MDL, and DT with EBD were used as base learners and LR as a meta-model. Table 4, and Table 5 illustrate the performance of our models in black-box and white-box scenarios, respectively. Each table shows the adversarial accuracy and robustness measures of the models against different adversarial attacks. Results show that the proposed models present high accuracy in adversarial environments and the lower robustness measure (RM) that confirms the effectiveness of our defense method. Looking at Table 4 and Table 2, we can see that EN-LR and EN-RF generally perform better compared to the other methods. For example, on the JSMA LR black attack to DT model, the EN-LR method has an Adv_Acc of 95.13% and an RM of 1.18%, while the discretization methods alone have Adv_Acc values ranging from 78.35% to 85.36% and RM values ranging from 6.89% to 8.81%. When Comparing Table 5 and Table 3, it is clear that the ensemble models generally outperform the individual models. For example, on JSMA white box attack to LR model, the EN-LR method has an Adv_Acc of 98.71% with an RM of 0.96%, while the discretization methods alone have Adv_Acc values ranging from 96.6% to 98.22% and RM values ranging from 1.66% to 3.19%. These results suggest that the ensemble models of EN-LR and EN-RF, which are built using single models trained on discretized input features as their base classifiers and LR and RF as their meta models, respectively, are more effective in improving the robustness of the models against adversarial attacks.

*Table 3 – proposed Model performance in black box scenario*

| Model | Black box attack | EN-LR | | EN-RF | |
|---|---|---|---|---|---|
| | | Adv_acc | RM | Adv_Acc | RM |
| **Decision Tree** | FGSM LR | 91.70% | 4.75% | 91.86% | 4.61% |
| | JSMA LR | 95.13% | 1.18% | 95.83% | 0.49% |
| | BIM LR | 93.07% | 3.32% | 91.02% | 5.48% |
| | JSMA FF | 91.15% | 5.32% | 92.54% | 3.90% |
| | FGSM FF | 94.50% | 1.84% | 92.70% | 3.74% |
| | BIM FF | 92.80% | 3.60% | 90.17% | 6.37% |
| **Random Forest** | FGSM LR | 97.77% | 1.94% | 97.34% | 2.38% |
| | JSMA LR | 99.36% | 0.34% | 99.59% | 0.12% |
| | BIM LR | 97.58% | 2.13% | 97.54% | 2.18% |
| | JSMA FF | 97.52% | 2.19% | 97.28% | 2.44% |
| | FGSM FF | 98.20% | 1.50% | 98.16% | 1.55% |
| | BIM FF | 96.58% | 3.13% | 96.55% | 3.17% |
| **LR** | JSMA FF | 95.95% | 3.73% | 96.71% | 2.89% |
| | FGSM FF | 97.89% | 1.79% | 96.49% | 3.11% |
| | BIM FF | 96.40% | 3.28% | 96.54% | 3.06% |
| **FF** | FGSM LR | 96.03% | 3.58% | 95.68% | 3.89% |
| | JSMA LR | 96.40% | 3.21% | 96.18% | 3.39% |
| | BIM LR | 96.39% | 3.22% | 95.98% | 3.59% |

*Table 5- proposed Model performance in white box scenario*

| | | EN-LR | | EN-RF | |
|---|---|---|---|---|---|
| **Model** | **Black box attack** | **Adv_acc** | **RM** | **Adv_Acc** | **RM** |
| **LR** | FGSM | 96.51% | 3.17% | 96.18% | 3.42% |
| | JSMA | 98.71% | 0.96% | 98.94% | 0.65% |
| | BIM | 96.39% | 3.29% | 95.13% | 4.48% |
| **FF** | JSMA | 94.62% | 5.00% | 94.28% | 5.29% |
| | FGSM | 93.50% | 6.12% | 93.42% | 6.16% |
| | BIM | 94.39% | 5.23% | 95.23% | 4.34% |

## 5 Conclusion

In this paper, we proposed discretization-based stack ensemble robustness to improve the robustness of machine learning-based IoT device identification models to adversarial attacks. we showed that not only discretization individually is effective and can increase model robustness to adversarial attacks. But also stack ensemble of different discretized models can significantly improve the model's robustness in both white and black box attack scenarios. This is the first step, and the techniques are likely to use in other areas, like intrusion detection and anomaly detection using different IoT datasets due to their general applicability and flexibility. Further analyses are left as future work.

## References

1. M. A. Al-Garadi, A. Mohamed, A. K. Al-Ali, X. Du, I. Ali, and M. Guizani, "A survey of machine and deep learning methods for internet of things (IoT) security," IEEE Communications Surveys & Tutorials, vol. 22, no. 3, pp. 1646-1685, 2020.
2. I. Cvitić, D. Peraković, M. Periša, and B. Gupta, "Ensemble machine learning approach for classification of IoT devices in smart home," International Journal of Machine Learning and Cybernetics, pp. 1-24, 2021.
3. N. Akhtar and A. Mian, "Threat of adversarial attacks on deep learning in computer vision: A survey," Ieee Access, vol. 6, 2018.
4. X. Yuan, P. He, Q. Zhu, and X. Li, "Adversarial examples: Attacks and defenses for deep learning," IEEE transactions on neural networks and learning systems, 2019.
5. A. Namvar, C. Thapa, S. S. Kanhere, and S. Camtepe, "Evaluating the Security of Machine Learning Based IoT Device Identification Systems Against Adversarial Examples," in Service-Oriented Computing: 19th International Conference, ICSOC 2021, Virtual Event, November 22–25, 2021, Proceedings 19, 2021: Springer, pp. 800-810.
6. P. Panda, I. Chakraborty, and K. Roy, "Discretization based solutions for secure machine learning against adversarial attacks," IEEE Access, vol. 7, pp. 70157-70168, 2019.

7. H. Mohanty, A. H. Roudsari, and A. H. Lashkari, "Robust stacking ensemble model for darknet traffic classification under adversarial settings," Computers & Security, vol. 120, p. 102830, 2022.

8. C. Szegedy et al., "Intriguing properties of neural networks," arXiv preprint 2013.

9. I. J. Goodfellow, J. Shlens, and C. Szegedy, "Explaining and harnessing adversarial examples (2014)," ICLR 2015, 2014.

10. W. Xu, D. Evans, and Y. Qi, "Feature squeezing: Detecting adversarial examples in deep neural networks," arXiv preprint arXiv:1704.01155, 2017.

11. K. Chen, H. Zhu, L. Yan, and J. Wang, "A survey on adversarial examples in deep learning," Journal on Big Data, vol. 2, no. 2, p. 71, 2020.

12. R. Feinman, R. R. Curtin, S. Shintre, and A. B. Gardner, "Detecting adversarial samples from artifacts," arXiv preprint arXiv:1703.00410, 2017.

13. A. Kurakin, I. Goodfellow, and S. Bengio, "Adversarial examples in the physical world," 2016.

14. N. Papernot, P. McDaniel, S. Jha, M. Fredrikson, Z. B. Celik, and A. Swami, "The limitations of deep learning in adversarial settings," in Security and Privacy (EuroS&P), 2016 IEEE European Symposium on, 2016: IEEE, pp. 372-387.

15. Z. Bao, Y. Lin, S. Zhang, Z. Li, and S. Mao, "Threat of adversarial attacks on DL-based IoT device identification," IEEE Internet of Things Journal, vol. 9, no. 11, pp. 9012-9024, 2021.

16. A. Singh and B. Sikdar, "Adversarial attack for deep learning based IoT appliance classification techniques," in 2021 IEEE 7th World Forum on Internet of Things (WF-IoT), 2021: IEEE, pp. 657-662.

17. Y. Meidan et al., "Detection of Unauthorized IoT Devices Using Machine Learning Techniques," arXiv preprint, 2017.

18. J. Kotak and Y. Elovici, "Adversarial Attacks Against IoT Identification Systems," IEEE Internet of Things Journal, 2022.

19. Z. Luo, S. Zhao, Z. Lu, Y. E. Sagduyu, and J. Xu, "Adversarial machine learning based partial-model attack in IoT," in Proceedings of the 2nd ACM Workshop on Wireless Security and Machine Learning, 2020, pp. 13-18.

20. J. Brownlee, Data preparation for machine learning: data cleaning, feature selection, and data transforms in Python. Machine Learning Mastery, 2020.

21. J. Buckman, A. Roy, C. Raffel, and I. Goodfellow, "Thermometer encoding: One hot way to resist adversarial examples," in International conference on learning representations, 2018.

22. S. Sharmin, N. Rathi, P. Panda, and K. Roy, "Inherent adversarial robustness of deep spiking neural networks: Effects of discrete input encoding and non-linear activations," in Computer Vision–ECCV 2020: 16th European Conference, Glasgow, UK, August 23–28, 2020, Proceedings, Part XXIX 16, 2020: Springer, pp. 399-414.

23. S. Lal et al., "Adversarial attack and defence through adversarial training and feature fusion for diabetic retinopathy recognition," Sensors, vol. 21, no. 11, p. 3922, 2021.

24. R. Feng, W.-c. Feng, and A. Prakash, "Essential Features: Content-Adaptive Pixel Discretization to Improve Model Robustness to Adaptive Adversarial Attacks," arXiv preprint arXiv:2012.01699, 2020.

25. K. Kireev, B. Kulynych, and C. Troncoso, "Adversarial Robustness for Tabular Data through Cost and Utility Awareness," arXiv preprint arXiv:2208.13058, 2022.

26. D. M. Maslove, T. Podchiyska, and H. J. Lowe, "Discretization of continuous features in clinical datasets," Journal of the American Medical Informatics Association, vol. 20, no. 3, pp. 544-553, 2013.

27. J. Chen, X. Wu, Y. Liang, and S. Jha, "Improving adversarial robustness by data-specific discretization," arXiv preprint arXiv:1805.07816, 2018.
28. T. H. A. Tran and M. Lara, "Influence of discretization granularity on learning classification models," in BNAIC/BeNeLearn 2022 Joint International Scientific Conferences on AI and Machine Learning, 2022.
29. A. Kurakin et al., "Adversarial attacks and defences competition," in The NIPS'17 Competition: Building Intelligent Systems, 2018: Springer, pp. 195-231.
30. F. Tramr, A. Kurakin, N. Papernot, I. Goodfellow, D. Boneh, and P. McDaniel, "Ensemble adversarial training: Attacks and defenses," in International Conference on Learning Representations, 2018, vol. 1, p. 2.
31. T. Pang, "Improving Adversarial Robustness via Promoting Ensemble Diversity."
32. E. Grefenstette, R. Stanforth, B. O'Donoghue, J. Uesato, G. Swirszcz, and P. Kohli, "Strength in Numbers: Trading-off Robustness and Computation via Adversarially-Trained Ensembles," 2018.
33. D. Li and Q. Li, "Adversarial Deep Ensemble: Evasion Attacks and Defenses for Malware Detection," arXiv preprint arXiv:2006.16545, 2020.
34. A. Dabouei, S. Soleymani, F. Taherkhani, J. Dawson, and N. M. Nasrabadi, "Exploiting joint robustness to adversarial perturbations," in Proceedings of the IEEE/CVF Conference on Computer Vision and Pattern Recognition, 2020, pp. 1122-1131.
35. A. Sivanathan et al., "Classifying IoT Devices in Smart Environments Using Network Traffic Characteristics," IEEE Transactions on Mobile Computing, 2018.
36. N. Papernot, P. McDaniel, and I. Goodfellow, "Transferability in machine learning: from phenomena to black-box attacks using adversarial samples," arXiv preprint 2016.
37. M.-I. Nicolae et al., "Adversarial Robustness Toolbox v1. 0.0," arXiv preprint arXiv:1807.01069, 2018.